\documentclass[11pt]{article}
\usepackage{geometry}
\usepackage{coling2020}
\usepackage{times}
\usepackage{url}
\usepackage{latexsym}
\usepackage{microtype}
\usepackage{subcaption}
\usepackage{caption}
\usepackage{multirow}
\usepackage{amsfonts}
\usepackage{graphicx}

\colingfinalcopy 

\title{GloVeInit at SemEval-2020 Task 1: Using GloVe Vector Initialization for Unsupervised Lexical Semantic Change Detection}

\author{Vaibhav Jain \\
  Delhi Technological University \\
  Delhi, India \\
  {\tt vaibhav29498@gmail.com} \\}

\date{}

\begin{document}
\maketitle
\begin{abstract}
This paper presents a vector initialization approach for the SemEval2020 Task 1: Unsupervised Lexical Semantic Change Detection. Given two corpora belonging to different time periods and a set of target words, this task requires us to classify whether a word gained or lost a sense over time (subtask 1) and to rank them on the basis of the changes in their word senses (subtask 2). The proposed approach is based on using Vector Initialization method to align GloVe embeddings. The idea is to consecutively train GloVe embeddings for both corpora, while using the first model to initialize the second one. This paper is based on the hypothesis that GloVe embeddings are more suited for the Vector Initialization method than SGNS embeddings. It presents an intuitive reasoning behind this hypothesis, and also talks about the impact of various factors and hyperparameters on the performance of the proposed approach. Our model ranks 13th and 10th among 33 teams in the two subtasks. The implementation has been shared publicly.\footnote{\url{github.com/vaibhav29498/GloVeInit}}
\end{abstract}

\section{Introduction and Background}
\blfootnote{This work is licensed under a Creative Commons Attribution 4.0 International License. License
details: \url{http://creativecommons.org/licenses/by/4.0/}.}
Lexical Semantic Change (LSC) Detection is an active research topic in the field of natural language processing, and has been applied for diachronic (across time) and synchronic (across domains) tasks \cite{schlechtweg2019wind}. Previously limited to manual "close-reading" approaches, the availability of large-scale corpora have allowed the use of computational methods for this task \cite{tahmasebi2018survey}. This topic has found various applications in various disciplines such as improving information retrieval from historical documents \cite{morsy2016accounting}, preventing cross-domain ambiguity in requirements elicitation interviews \cite{jain2020cross}, and studying the impact of societal and cultural changes on word meanings and usage \cite{tahmasebi2017uses}.

LSC detection involves the use of two corpora $C_1$ and $C_2$ which, in the diachronic case, belong to different time periods $t_1$ and $t_2$ respectively. The various approaches found in the literature usually involve the construction of a word embedding space specific to each corpus. The embeddings can be constructed through count-based methods such as Positive Pointwise Mutual Information and Random Indexing, or predictive methods such as Skip-Gram with Negative Sampling (SGNS). Most of these models are stochastic in nature which means that separately trained embedding models live in their own space. In order to project them onto a \emph{unified space}, alignment techniques such as vector initialization and orthogonal Procrustes are used. The LSC of a word is then quantitatively determined by measuring the contextual dissimilarity between the word's representations \cite{tahmasebi2018survey}.

Recently, there have been efforts to evaluate these various methods by comparing their results with manually-annotated data \cite{schlechtweg2019wind,ahmad2020shared}. The SemEval-2020 Task 1 is one such effort which is based on LSC detection in corpora of English, German, Latin, and Swedish languages \cite{schlechtweg2020semeval}. Its aim is to provide an evaluation framework for unsupervised LSC detection systems by comparing their results against a ground truth, as annotated by native speakers or scholars. It consists of two subtasks: \textit{Given two corpora $C_1$ and $C_2$ (for time periods $t_1$ and $t_2$), for a set of target words,}
\begin{enumerate}
    \item \textit{decide which words lost or gained senses between $t_1$ and $t_2$, and which ones did not; as annotated by human judges.}
    \item \textit{rank them according to their degree of LSC between $t_1$ and $t_2$ as annotated by human judges. A higher rank means stronger change.}
\end{enumerate}

\section{System Overview}

\subsection{Vector Initialization}
\label{sec:vi}
The Vector Initialization (VI) alignment method, which was first used by \newcite{kim2014temporal}, involves the embedding space for $t_1$ to be trained independently on $C_1$. It is then used to initialize the embedding space for $t_2$ which is then subsequently trained from $C_2$. The underlying idea is that the embedding for a word $w$ will get considerably updated if it is used within different contexts in $C_1$ and $C_2$, otherwise it will receive only a slight update. A study by \newcite{schlechtweg2019wind}, which applied the VI method on SGNS embeddings, found it to perform significantly weaker on LSC detection tasks than other methods such as orthogonal Procrustes (OP). They attributed this to the sensitivity of the VI method to the frequency of a word in $C_2$. The high frequency of a word in $C_2$ will result into the word's embedding getting frequent updates away from its initial state. This leads to a large divergence between the word's representations even if it does not undergo any LSC from $t_1$ to $t_2$.

In a recent shared task on LSC detection in the German language, a modified VI approach achieved the third-best result and outperformed certain OP approaches \cite{ahmad2020shared}. Instead of only initializing on the word embeddings obtained from $C_1$, the modified approach initializes on the complete SGNS model which includes the hidden layer. However, it also theoretically suffers from sensitivity to high frequency.

\subsection{Comparison of GloVe and SGNS}

GloVe (Global Vectors) and SGNS are unsupervised algorithms for obtaining word embedding spaces \cite{pennington2014glove,mikolov2013distributed}. GloVe makes use of co-occurrence matrix $X$; its $(i, j)$ entry, $X_{ij}$ is the number of times the word $w_j$ appears in the context of the word $w_i$ (as defined by the window-size $L$). It trains the word embeddings by minimizing the cost function

\begin{equation}
    J =  \sum^V_{i = 1} \sum^V_{j = 1; j \neq i} f(X_{ij}) (u^T_j v_i - \log X_{ij})^2,
\end{equation}

where $V$ is the vocabulary size and $u, v \in \mathbb{R}^D$ are the word and context word vectors respectively. The final word embeddings can be obtained by summing or averaging the two.

On the other hand, SGNS is a predictive model which relies on a shallow two-layer neural network which, given a word, predicts the set of its context words. To avoid the expensive softmax function in the training objective, negative sampling is used by drawing a few negative samples from a noise distribution.

An important distinction between GloVe and SGNS from this paper's point of view is the number of updates that are made to a word embedding during training. In a single epoch, the number of updates to the SGNS embedding of a word is roughly equal to the number of words that appear in its context throughout the corpus. Hence, the number of updates is proportional to the frequency of the word which has an upper-bound of the total number of words in the corpus. However, this relationship is not exactly linear because of downsampling of frequent words and negative sampling. In the GloVe model, the number of updates received in a single epoch is equal to the number of \emph{distinct} context words. This number is limited by an upper-bound of vocabulary size $V$, which is usually much less than the total number of words in the corpus.

\begin{figure}[ht]
\centering
\begin{subfigure}[b]{.48\textwidth}
  \includegraphics[width=\linewidth,height=\linewidth]{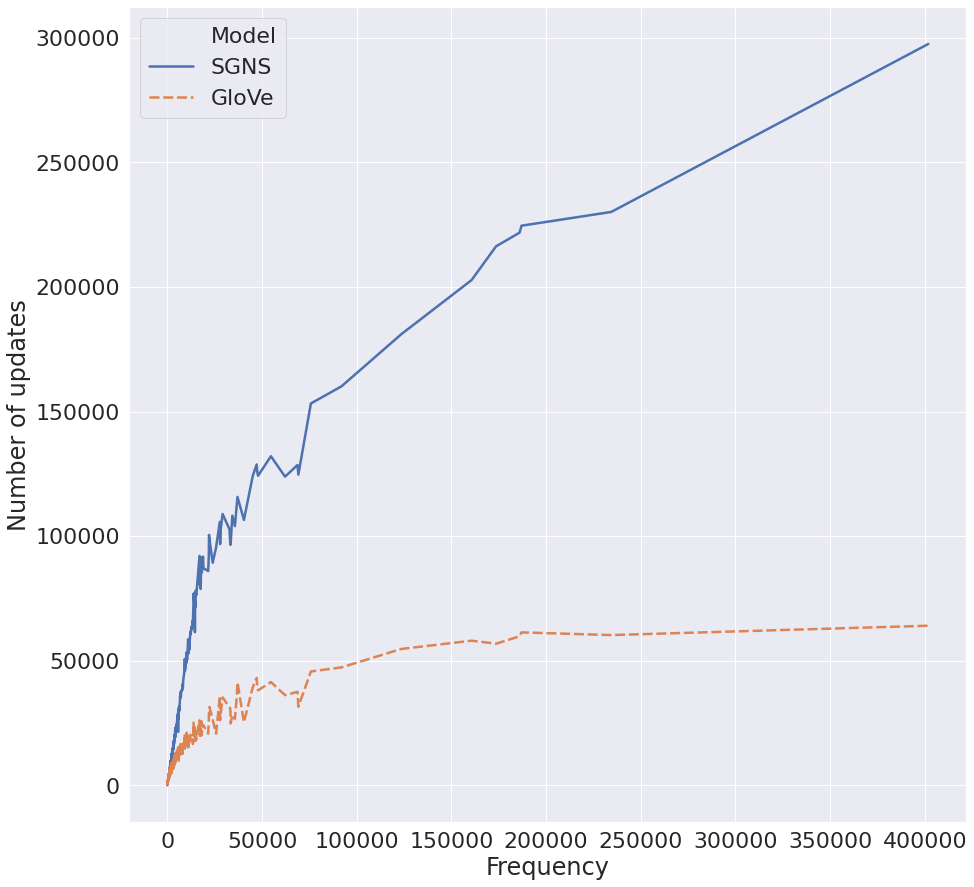}
  \centering
  \caption{CCOHA 1960-2010 (English)}
  \label{fig:english}
\end{subfigure}
\begin{subfigure}[b]{.49\textwidth}
  \includegraphics[width=\linewidth,height=\linewidth]{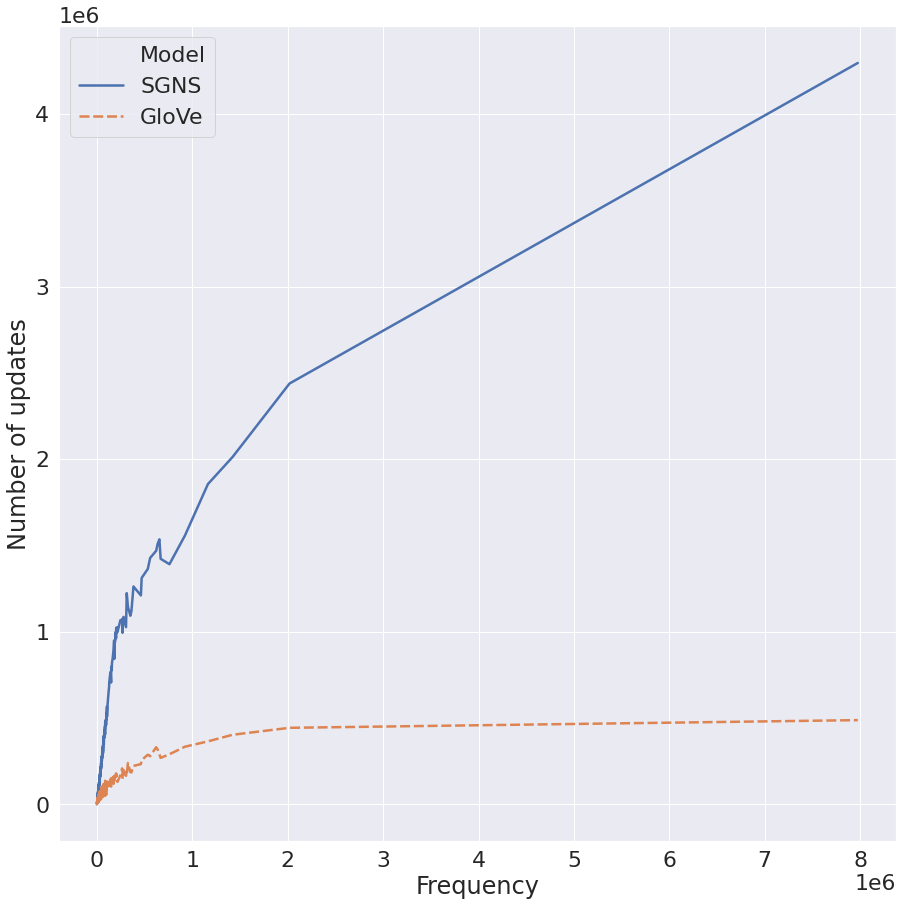}
  \centering
  \caption{BZ and ND 1946-1990 (German)}
  \label{fig:german}
\end{subfigure}

\vspace{\baselineskip}

\begin{subfigure}[b]{.48\textwidth}
  \includegraphics[width=\linewidth,height=\linewidth]{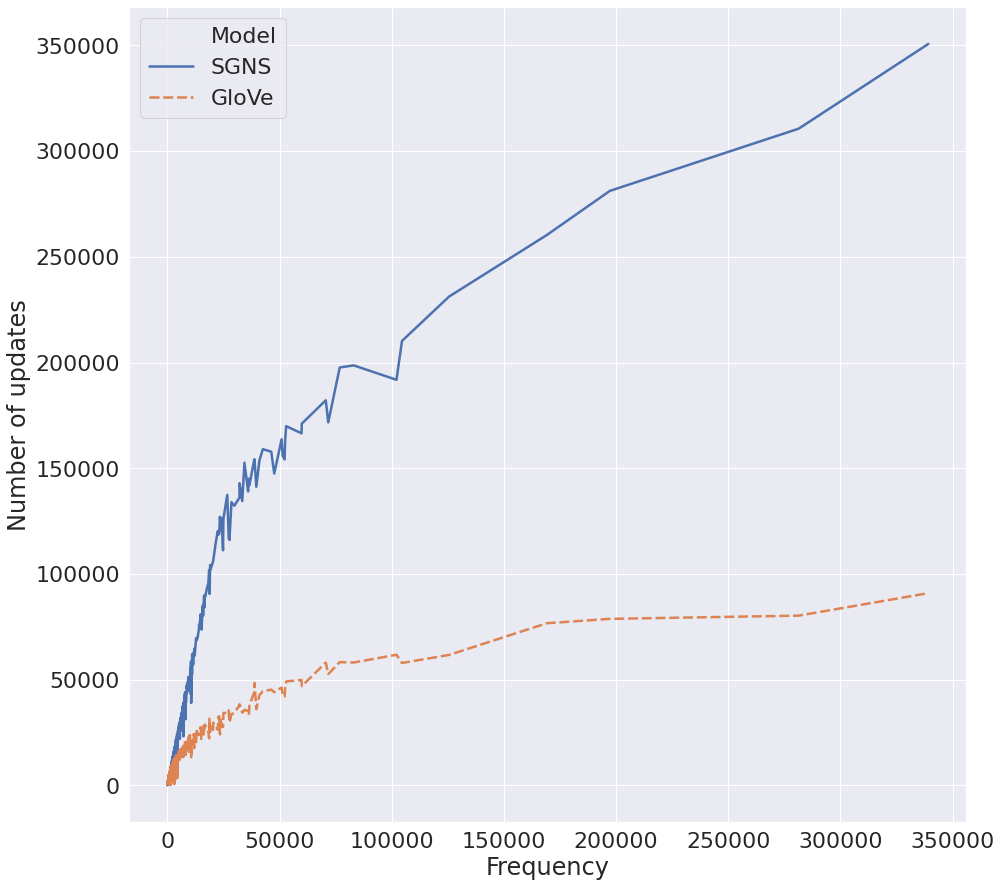}
  \centering
  \caption{LatinISE 0-2000 (Latin)}
  \label{fig:latin}
\end{subfigure}
\begin{subfigure}[b]{.49\textwidth}
  \includegraphics[width=\linewidth,height=\linewidth]{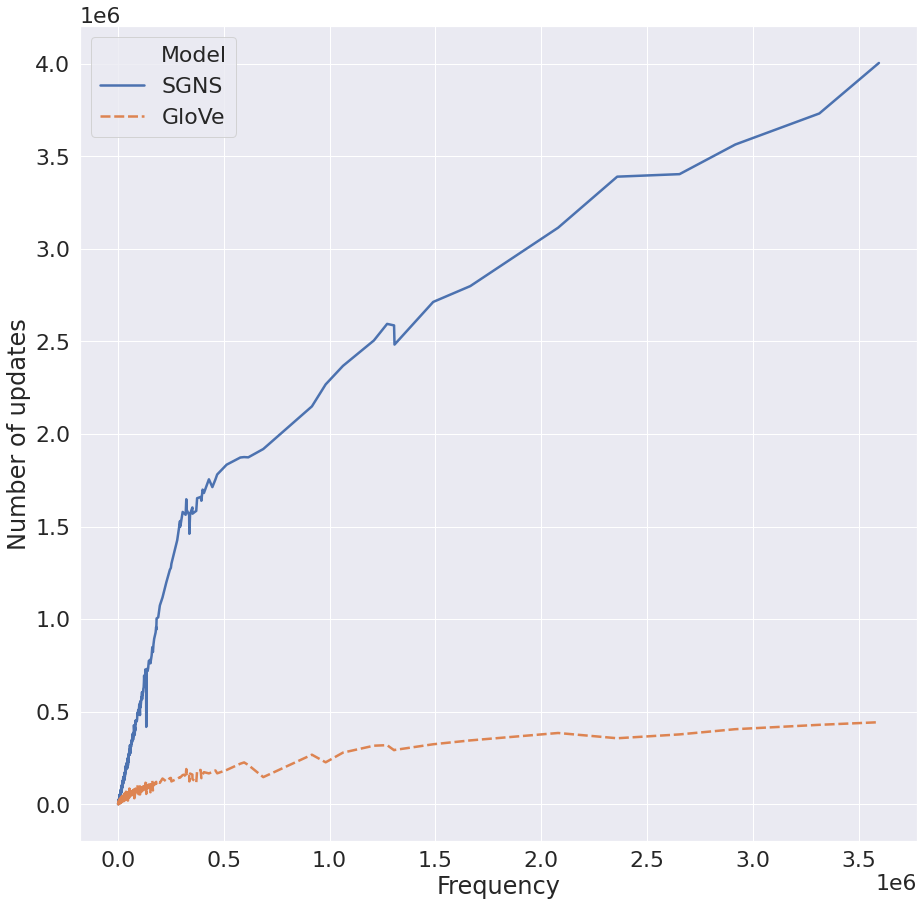}
  \centering
  \caption{KubHist 1895-1903 (Swedish)}
  \label{fig:swedish}
\end{subfigure}
\caption{Plot of frequency of a word (x-axis) vs the number of updates to its embedding (y-axis). The subcaptions include the name of the corpora $C_2$, the corresponding time-period $t_2$, and its language.}
\label{fig:lineplots}
\end{figure}

As discussed in Section~\ref{sec:vi}, the VI method can falsely give a high LSC score to words with high frequency. Our hypothesis is that the GloVe model is more suited for the VI method due to its lesser sensitivity to high frequency. To give an indication of the extent of this difference, the relationship between a word's frequency and the number of updates to its embedding is depicted in Figure \ref{fig:lineplots} for both GloVe and SGNS models trained on the $C_2$ corpora of all the four languages. These results are based on the official GloVe implementation\footnote{\url{github.com/stanfordnlp/GloVe}} and the \texttt{gensim}\footnote{\url{radimrehurek.com/gensim}} implementation of SGNS. The models were trained for a single epoch and only words with a frequency of greater than or equal to 5 were considered. It is clear that the proportion of updates for frequent and rare words is comparatively more balanced in the case of GloVe. On the other hand, there is a high rate of growth in the number of updates with respect to the frequency in the case of SGNS, which indicates that it is biased towards estimating a high LSC for frequent words.

\section{Results and Experimental Setup}

The dataset provided by the competition organizers consists of the corpora $C_1$ and $C_2$, and the list of target words for four languages: English, German, Latin, and Swedish \cite{schlechtweg2020semeval}. All of the corpora were already prepossessed: they were in tokenized and lemmatized form with punctuation marks and one-word sentences removed.

Subtask 1 is concerned with identifying the loss or gain of one or more word sense(s), where subtask 2 tests a model's ability to detect fine-grained changes in the two sense frequency distributions. For example, consider the word \emph{cell} which historically referred to either a chamber or the smallest unit of an organism. However, the relative usages of both these senses have decreased with time, with mobile phone being the predominant meaning now. For subtask 2, the ground-truth ranking of the target words is determined by calculating the Jensen-Shannon divergence between their normalized sense frequency distributions from $t_1$ and $t_2$ \cite{donoso2017dialectometric}. A submission is scored by its Spearman's rank-order correlation coefficient against the ground-truth ranking.

We used cosine distance as the metric to calculate the distance between a word's vectors in diachronic spaces, and every word with a distance of more than 0.55 (for German) or 0.45 (for other languages) were classified to have gained of lost word sense(s).

In all of our experiments, only words having a frequency of at least 5 were considered. Our best-performing solution during the evaluation phase achieved an accuracy of 60\% in subtask 1 and a score of 0.352. It was obtained by training word embeddings with a dimensionality of 50 on both $C_1$ and $C_2$ for 60 epochs each with a window-size of 10. The subtask-2 score was close to but less than the modified SGNS-based VI approach discussed in Section \ref{sec:vi}, which was proposed by the team \emph{IMS} and received a score of 0.372. We ranked 13th and 10th out of 33 participants in the two subtasks respectively.

Further experiments on subtask-2 were conducted after the competition for analyzing the impact of hyperparameters like window-size $L$ and embedding dimensionality $d$ and are reported in Table \ref{table:exp}. Number of training epochs was limited to 20 for models with window-size 10 because of their high computational requirements. The results suggest that training the models for higher number of epochs can produce better results if the embedding dimensionality is high, but can backfire in the opposite case. Using a larger window-size improves the average result in most cases.

\begin{table}[ht]
\centering
\begin{tabular}{|c|c|c|c|c|c|c|c|}
\hline \multirow{2}{*}{\bf \em L} & \multirow{2}{*}{\bf Epochs} & \multirow{2}{*}{\bf \em d} & \multicolumn{5}{c|}{\bf Scores} \\
\cline{4-8}
& & & \bf English & \bf German & \bf Latin & \bf Swedish & \bf Average \\
\hline
\multirow{10}{*}{5} & \multirow{5}{*}{20} & 5 & 0.012 & 0.366 & 0.391 & 0.15 & 0.23 \\
& & 10 & -0.144 & 0.362 & 0.394 & 0.114 & 0.182 \\
& & 20 & 0.193 & 0.402 & 0.368 & 0.244 & 0.302 \\
& & 50 & \bf 0.278 & 0.446 & 0.315 & 0.229 & 0.317 \\
& & 100 & 0.226 & 0.312 & 0.254 & 0.186 & 0.244 \\
\cline{2-8}
& \multirow{5}{*}{60} & 5 & -0.018 & 0.354 & 0.432 & 0.121 & 0.222 \\
& & 10 & -0.095 & 0.283 & 0.379 & 0.108 & 0.169 \\
& & 20 & 0.212 & 0.41 & 0.377 & 0.324 & 0.331 \\
& & 50 & 0.228 & 0.464 & 0.329 & \bf 0.366 & \bf 0.347 \\
& & 100 & 0.269 & 0.312 & 0.276 & 0.309 & 0.291 \\
\hline
\multirow{5}{*}{10} & \multirow{5}{*}{20} & 5 & 0.011 & 0.334 & \bf 0.485 & 0.218 & 0.262 \\
& & 10 & -0.036 & 0.351 & 0.409 & 0.146 & 0.218 \\
& & 20 & 0.144 & \bf 0.479 & 0.397 & 0.187 & 0.302 \\
& & 50 & 0.198 & 0.477 & 0.275 & 0.267 & 0.304 \\
& & 100 & 0.199 & 0.349 & 0.257 & 0.227 & 0.258 \\
\hline
\end{tabular}
\caption{Results of the post-evaluation experiments}
\label{table:exp}
\end{table}

\subsection{Error Analysis}
\begin{figure}[ht]
    \centering
    \includegraphics[width=\columnwidth]{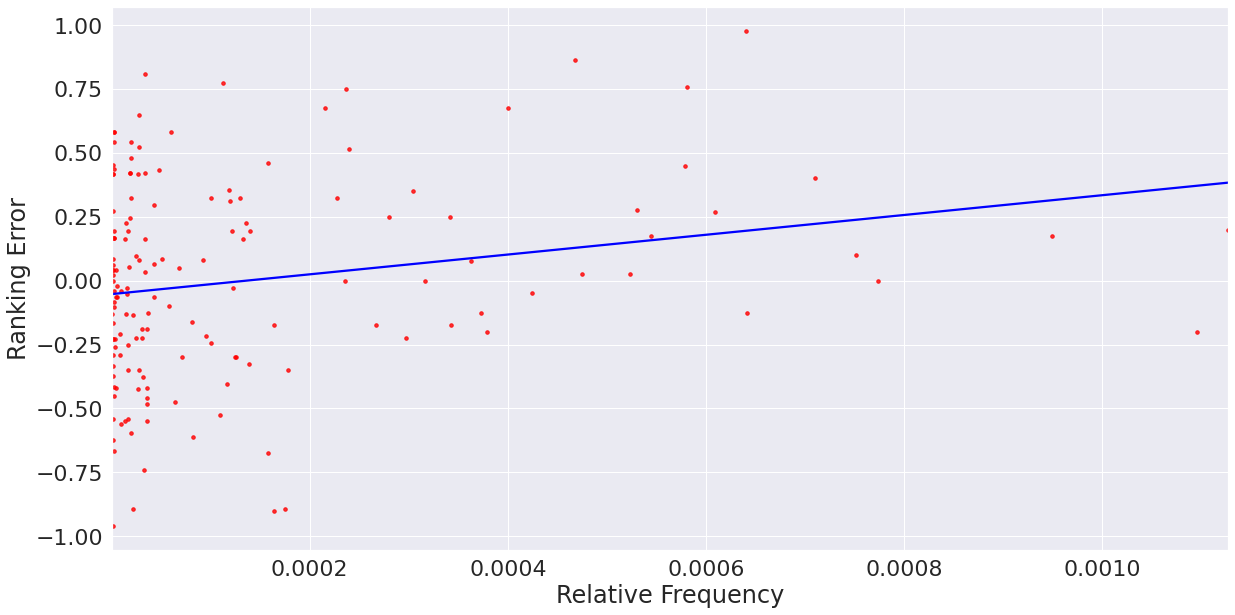}
    \caption{Regression plot between the frequency of the target words and their ranking error}
    \label{fig:regplot}
\end{figure}

We defined the ranking error for each target word as the difference between its predicted rank and its true rank divided by the total number of target words. It lies in the range $(-1, 1)$ and has an ideal value of zero. A positive value indicates that the model overestimated the LSC for a word, and a negative value indicates otherwise. A regression plot between the relative frequency of the target words and their ranking error as per our best performing model is depicted in Figure \ref{fig:regplot}. A word's relative frequency is defined as its frequency divided by the total frequency of all words in the corpus.

There is a statistically significant correlation between the ranking error and the relative frequency, with the result of the Spearman's rank-order correlation test being $\rho = 0.186$ and $p = 0.020$. This indicates that the GloVe model's relative insensitivity to high frequency can lead to it assigning a rank lower than the true rank to such words. We believe that the frequency of the target words is not large enough for SGNS' sensitivity to high frequency to have a major impact on the results. This explains our model's inferior performance as compared to SGNS-based model in this task.

\section{Conclusion and Future Work}
In this paper, we reported our work in the SemEval2020 Task 1. We proposed a GloVe-based VI approach which achieved the 11th rank out of 32 participating teams. We gave a theoretical reasoning behind why GloVe is less sensitive towards high frequency than SGNS and thus more suited for VI method, and empirically showed the magnitude of the difference between the number of updates to word embeddings in the two models. We believe that despite a lower-than-expected performance in this competition, our work presents a good case for the suitability of GloVe model when corpora of larger size are involved. However, proving this quantitatively is a challenging task because of the limitations associated with manual annotation-based evaluation.

Planned future work includes the study of how techniques such as dimension-wise mean-centering and length normalization, which have proved beneficial in OP-based approaches, can be applied for VI method.

\section*{Acknowledgments}

We would like to thank the task organizers for preparing the dataset and organizing the competition, Jens Kaiser (member of the team \emph{IMS}) from University of Stuttgart for sharing the results from his work on the SGNS-based VI alignment method, and the anonymous reviewers for their detailed and helpful feedback.

\bibliographystyle{coling}
\bibliography{semeval2020}

\end{document}